\documentclass[arxiv]{melba}

\usepackage{amsmath,amsfonts}

\usepackage{xspace}
\usepackage{subcaption}
\usepackage[nolist]{acronym}
\usepackage[capitalise]{cleveref}
\usepackage{siunitx}
\newcommand{\ie}{i.\,e.\xspace}

\newcommand{\plasm}{\emph{Plasmodium}\xspace}
\newcommand{\plasmf}{\emph{P.\,falciparum}\xspace}
\newcommand{\plasmv}{\emph{P.\,vivax}\xspace}
\newcommand{\plasmm}{\emph{P.\,malariae}\xspace}
\newcommand{\plasmo}{\emph{P.\,ovale}\xspace}
\newcommand{\plasmk}{\emph{P.\,knowlesi}\xspace}
\newcommand{\polys}{NIH$_\emph{polys}$\xspace}
\newcommand{\points}{NIH$_\emph{points}$\xspace}
\newcommand{\mira}{MIRA$_\emph{boxes}$\xspace}
\newcommand{\rcnn}{Faster R-CNN\xspace}

\melbaid{2025:022}  
\doi{10.59275/j.melba.2025-3ecf}
\melbaauthors{Wilm et al.}  
\email{fwilm@mira.vision}
\volume{3}
\firstpageno{1}  
\melbayear{2025}  
\datesubmitted{yyyy-m1-d1}  
\datepublished{yyyy-m2-d2}  

\melbaspecialissue{MICCAI Open Data special issue}
\melbaspecialissueeditors{Martijn Starmans, Apostolia Tsirikoglou, Lidia Garrucho Moras, Kaouther Mouheb}

\ShortHeadings{Instance-Level Plasmodium Falciparum Detection}{Wilm et al.}

\title{A COCO-Formatted Instance-Level Dataset for Plasmodium Falciparum Detection in Giemsa-Stained Blood Smears}

\author{
	\firstname Frauke \surname Wilm\aff{1,2}\orcid{0000-0002-9065-0554},
	\name Luis Carlos Rivera Monroy\aff{1, 2}\orcid{0000-0001-8232-8920},
    \name Mathias Öttl\aff{1, 2}\orcid{0009-0004-6521-5843},
    \name Lukas Mürdter\aff{1},
    \name Leonid Mill\aff{1, 2}\orcid{0000-0001-6449-7309},
    \name Andreas Maier\aff{2}\orcid{0000-0002-9550-5284}
}
\affiliations{
	\num 1 \addr MIRA Vision Microscopy GmbH, 73037 Göppingen, Germany \\
	\num 2 \addr Pattern Recognition Lab, Department of Computer Science,
   Friedrich-Alexander-Universität (FAU) Erlangen-Nürnberg,
   Erlangen, Germany
}

\abstract{
Accurate detection of \emph{Plasmodium falciparum} in Giemsa-stained blood smears is an essential component of reliable malaria diagnosis, especially in developing countries. Deep learning-based object detection methods have demonstrated strong potential for automated Malaria diagnosis, but their adoption is limited by the scarcity of datasets with detailed instance-level annotations. In this work, we present an enhanced version of the publicly available NIH malaria dataset, with detailed bounding box annotations in COCO format to support object detection training. We validated the revised annotations by training a Faster R-CNN model to detect infected and non-infected red blood cells, as well as white blood cells. Cross-validation on the original dataset yielded F1\,scores of up to 0.88 for infected cell detection. These results underscore the importance of annotation volume and consistency, and demonstrate that automated annotation refinement combined with targeted manual correction can produce training data of sufficient quality for robust detection performance. The updated annotations set is publicly available via Zenodo: \url{https://doi.org/10.5281/zenodo.17514694}.
}


\begin{document}

\twocolumn[\maketitle]

\section{Background}
\enluminure{M}{alaria} is a tropical disease caused by protozoan parasites of the genus \plasm, which infect red blood cells and are primarily transmitted through the bites of female Anopheles mosquitoes. In humans, the disease is mainly associated with four species: \plasmf, \plasmv, \plasmm, and \plasmo. In recent years, malaria has also been increasingly transmitted by a fifth species, \ie, \plasmk. Among these five species, \plasmf and \plasmv are the most prevalent, and \plasmf is responsible for the majority of malaria-related deaths~\citep{world2024guidelines}.  

Most malaria infections are reported in tropical and subtropical regions, affecting populations in low-income countries with limited access to healthcare. Although modern treatments can effectively cure malaria, early diagnosis remains critical and delays in detection are a major contributing factor to malaria-related mortality~\citep{sultani2022towards}. For parasitological diagnosis of malaria, microscopic examination of thick and thin blood smear images is routinely performed. In addition to identifying the \plasm species, light microscopy allows parasite quantification and monitoring therapy response. Therefore, it is often preferred over molecular testing~\citep{world2024guidelines}. Nonetheless, the parasitological assessment of blood smear images requires a high level of expertise, and trained personnel might be scarce in low-resource countries or rural areas~\citep{poostchi2018image}.      

Recently, machine learning-based approaches for analyzing digitized blood smear images have demonstrated promising results in parasitemia quantification~\citep{poostchi2018image}. However, these methods typically rely on large, well-annotated datasets for effective training, making publicly available resources particularly valuable. Most existing work focuses on classifying individual cell patches as infected or non-infected~\citep{kassim2020clustering}, which requires the prior extraction of single-cell crops. This step can be challenging in densely populated blood smear images and limits the applicability of such approaches in real-world diagnostic workflows, where direct localization and accurate quantification of infected cells are essential. In contrast to patch-based classification approaches, object detection architectures require datasets with detailed instance-level annotations, typically in the form of labeled bounding boxes. However, acquiring such detailed annotations is labor-intensive and time-consuming, which limits their availability. The \acs{nih} dataset, comprising 965 images, is one of the largest publicly available resources for \plasmf detection. However, only \num{165} of these images include detailed polygon-based annotations, while the remaining \num{800} are limited to point annotations marking cell centers. This sparsity limits their suitability for training deep learning-based object detection models, which typically require bounding box annotations. 

In this work, we present a revised version of the \acs{nih} dataset with enhanced annotations. Using the Cellpose framework~\citep{pachitariu2022cellpose} and manual label correction, we converted the original point annotations into bounding box labels, which are better suited for object detection. To validate the quality of the revised dataset, we trained a \rcnn~\citep{ren2015faster} for parasite detection, achieving an F1\,score of up to \num{0.88} for infected cell identification. The updated annotation set is publicly available via Zenodo: \url{https://doi.org/10.5281/zenodo.17514694}.

\section{Methods}
For our experiments, we generated new bounding box annotations for the \acs{nih} dataset, which contains Giemsa-stained, thin blood smear images of \plasmf. We conducted a technical validation of these annotations by training a deep learning-based object detector to identify three cell types: non-infected red blood cells, infected red blood cells, and white blood cells.

\subsection{Data Details}
The \acs{nih} dataset~\citep{kassim2020clustering} is a thin-smear malaria image dataset acquired at Chittagong Medical College Hospital in Bangladesh and published by the National Library of Medicine, \acf{nih}, Bethesda, MD, USA. It comprises Giemsa-stained, thin blood smear images from 193 patients (148 infected and 45 uninfected), with five images per patient. Each image was captured using a microscope-mounted smartphone camera at a resolution of \num{5312}$\,\times\,$\num{2988} (width$\,\times\,$height) pixels. Annotations cover three classes: non-infected red blood cells, infected red blood cells, and white blood cells. Of the \num{965} total images, \num{165} include detailed polygon-based annotations, while the remaining \num{800} provide only point annotations marking cell centers. \Cref{tab:datasets} summarizes these subsets, hereafter referred to as \polys and \points, respectively. \Cref{fig:poly,fig:points} show example regions of interest with contour and point annotations, corresponding to the \polys and \points subsets.

\begin{table}[ht]
\centering
\caption{\label{tab:datasets} Overview of the \acs{nih} dataset subsets. \polys includes detailed polygon-based labels, whereas \points was annotated with point markers indicating cell centers. \mira comprises revised labels for the \points dataset with detailed bounding box annotations.}
\resizebox{\linewidth}{!}{
\begin{tabular}{lS[table-format=5.0]S[table-format=5.0]S[table-format=5.0]}
\hfill & \text{\polys} & \text{\points} & \text{\mira} \\
\hline
patients & 33 & 160 & 160\\
no. of images & 165 & 800 & 800 \\
annotations & \text{contours} & \text{points} & \text{boxes} \\ 
no. of annotations & & & \\ 
\hspace{.25cm} non-infected & 33071 & 155640 & 155201\\
\hspace{.25cm} infected & 1142 & 6810 & 6805\\
\hspace{.25cm} white blood cell & 51  & 220 &  220\\
\hspace{.25cm} ambiguous & \hfill \text{-}  & \hfill \text{-} &  19592\\
\end{tabular}}
\end{table}

\subsection{Annotation Revision}
\label{sec:annos}
To enable the use of the \acs{nih} dataset for training object detection models, we converted the point annotations into detailed bounding-box annotations. For this, we first detected cell instances using Cellpose 2~\citep{pachitariu2022cellpose}, an open-source framework designed for robust, generalizable segmentation. Trained with a diverse dataset of more than \num{70000} cells, Cellpose offers strong performance across a wide range of cell types and imaging modalities, making it well suited for segmenting Giemsa-stained blood smear images.   

\begin{figure*}[ht]
\centering
\begin{subfigure}[b]{0.33\linewidth}
\centering
\caption{\label{fig:poly} Sample from \polys}
\includegraphics[height=4cm]{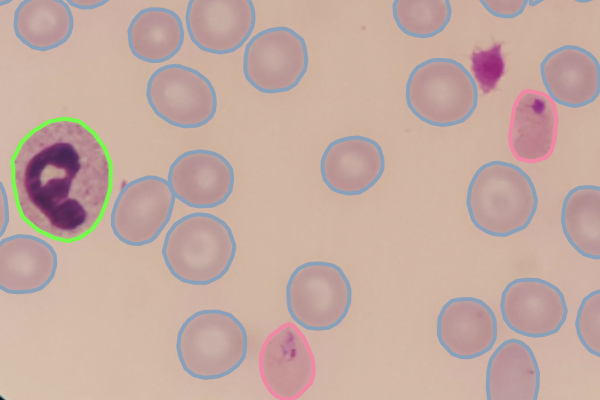}
\end{subfigure}%
\begin{subfigure}[b]{0.33\linewidth}
\centering
\caption{\label{fig:points} Sample from \points}
\includegraphics[height=4cm]{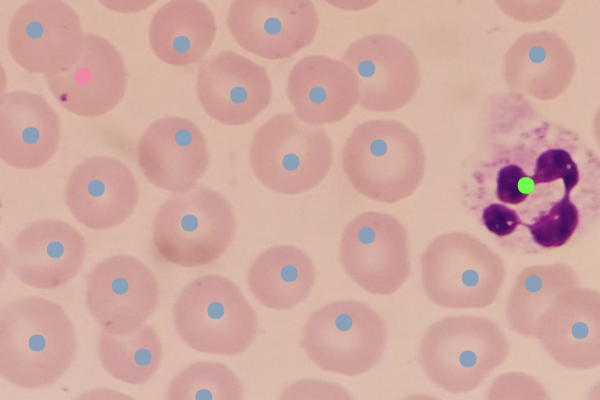}
\end{subfigure}%
\begin{subfigure}[b]{0.33\linewidth}
\centering
\caption{\label{fig:boxes} Bounding box annotations}
\includegraphics[height=4cm]{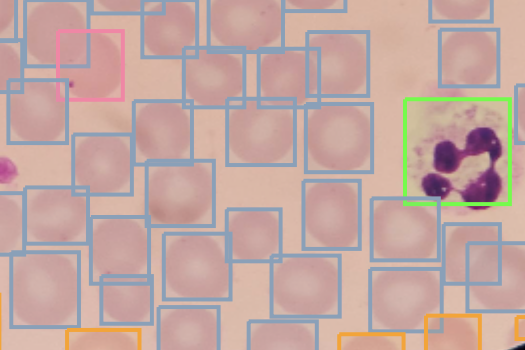}
\end{subfigure}
\caption{\label{fig:annos} Different annotation types provided by the \ac{nih} dataset. (a): contour annotations, (b): point-only annotations, (c) bounding box annotations created with Cellpose~\citep{pachitariu2022cellpose}. Blue: non-infected red blood cells, pink: infected cells, green: white blood cells, orange: ambiguous cells.}
\end{figure*}

\begin{figure}[ht]
    \centering
    \includegraphics[width=\linewidth]{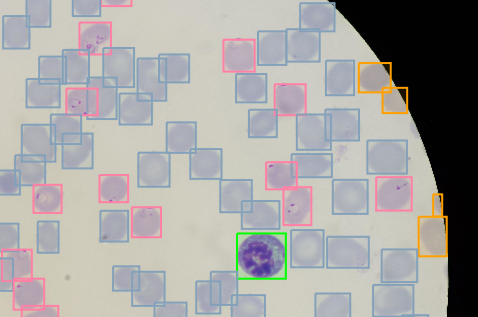}
    \caption{During label cleaning, non-annotated cells at the border of the field of view were labeled as \emph{ambiguous} (orange). Blue: non-infected red blood cells, pink: infected cells, green: white blood cells.}
    \label{fig:ambiguous}
\end{figure}

Following cell instance segmentation, we assigned labels to detected cells by overlaying the original point annotations. If a point annotation fell within a predicted bounding box, that box was assigned the corresponding cell class. However, Cellpose occasionally detected cells, which were not annotated in the original dataset. These were often partially visible cells at the edge of the field of view. In the updated annotation set, these detections were labeled as \emph{ambiguous}. \Cref{fig:ambiguous} shows an example with ambiguous cells at the border of the field of view. Overall, the updated annotations comprise \num{19592} ambiguous cells, which makes up around \SI{10}{\percent} of the original \points subset. 

Due to its reliance on an average cell size, Cellpose sometimes fragmented larger cells and particularly white blood cells into multiple instances. To address this, we manually reviewed and merged these fragmented detections. Additionally, Cellpose occasionally misclassified artifacts or blood platelets as cells. These false positives were also removed during manual post-processing.  \Cref{fig:boxes} shows a representative region of interest after bounding box detection, with ambiguous cells highlighted in orange, and the last column of \cref{tab:datasets} summarizes the number of cell instances after this annotation revision.

\section{Technical Validation}
To validate the revised annotations, we trained a \rcnn model~\citep{ren2015faster} to detect three cell classes: non-infected red blood cells, infected red blood cells, and white blood cells. We conducted cross-validation experiments by training the model on either the \polys or the revised \mira subset and evaluating its detection performance on the other, respectively.

\subsection{Implementation Details}
We employed a \rcnn model~\citep{ren2015faster} with a ResNet34~\citep{he2016deep} backbone, pretrained on ImageNet~\citep{russakovsky2015imagenet}. The datasets were split into \SI{70}{\percent} for training and \SI{30}{\percent} for validation. On the \polys dataset, the model was trained for \num{1000} epochs using a cosine annealing learning rate schedule with linear warm-up over the first 50 epochs and a maximum learning rate of \num{e-4}. For the \mira dataset, the training time was lowered to \num{200} epochs, to match the almost five-fold size of the data subset. For optimization, the Adam optimizer and standard \rcnn loss functions were used. Training patches of \num{1280}$\,\times\,$\num{960} pixels were sampled from the original \num{5312}$\,\times\,$\num{2988} pixel images. This resolution was chosen to match the 4:3 aspect ratio typical of microscopy images, while ensuring that each patch contained a sufficient number of cells for effective training. The patches were then downscaled by a factor of 2 to a final size of \num{640}$\,\times\,$\num{480} pixels, enabling a batch size of \num{32} without exceeding memory constraints. To address class imbalance, we applied a custom patch sampling strategy that over-sampled regions containing underrepresented classes, such as white blood cells. Model performance was monitored using the \ac{map} on the validation set, and the final model was selected based on the best validation \ac{map}. 

For inference on the full-resolution \num{5312}$\,\times\,$\num{2988} pixel images, we used the SAHI framework~\citep{akyon2021sahi,akyon2022sahi}, which performs sliding-window predictions and applies \ac{nms} to eliminate duplicate detections across overlapping patches. As a post-processing step, we removed all predicted bounding boxes with an area smaller than \num{2500} pixels or larger than \num{140000} pixels. These thresholds were determined based on the minimum and maximum annotation sizes observed in the original \ac{nih} dataset.

Training was performed on an NVIDIA A100 GPU. Experiments were implemented using the torchvision Faster R-CNN model, with PyTorch Lightning~\citep{lightning} for streamlined training and Hydra~\citep{Yadan2019Hydra} for configuration management.

\subsection{Evaluation}
\label{sec:eval}
For evaluation, we computed class-wise F1\,scores from the instance-level confusion matrices. Cells that were detected by Cellpose but not labeled by human annotators (\ie, ambiguous cells) were excluded from the evaluation. Annotated cells that were not detected by the model were considered \ac{fn-det}, while model predictions that were not annotated and not labeled as ambiguous were considered \ac{fp-det}. The class-wise F1\,score for class $c$ was computed as:

\begin{align}
F1(c) &= 2 \cdot \frac{\text{Prec}(c) \cdot \text{Rec}(c)}{\text{Prec}(c) + \text{Rec}(c)} \enspace \text{, with} \label{eq:f1}
\end{align}

\begin{align}
\nonumber \text{Prec}(c) &= \frac{TP(c)}{TP(c) + FP_{\text{cls}}(c) + FP_{\text{det}}(c)}  \\ \label{eq:prec} &= \frac{M_{cc}}{\sum_{i=1}^{N+1} M_{ic}} \enspace \text{, and}
\end{align}

\begin{align}
\nonumber \text{Rec}(c)  &= \frac{TP(c)}{TP(c) + FN_{\text{cls}}(c) + FN_{\text{det}}(c)} \\ \label{eq:rec} &= \frac{M_{cc}}{\sum_{i=1}^{N+1} M_{ci}} \enspace \text{.}
\end{align}

\noindent Here, $M_{ij}$ denotes the element in the $i$-th row and $j$-th column of the confusion matrix, \ie, the number of cells labeled as class $i$ and predicted as class $j$. $N$ is the number of cell classes, and the $(N+1)$-th row and column represent false positive (\ac{fp-det}) and false negative (\ac{fn-det}) detections, respectively.

\subsection{Results}
\begin{figure}[ht]
\centering
\begin{subfigure}[b]{\linewidth}
\centering
\caption{\label{fig:cm_points} train: \polys, test: \mira}
\includegraphics[width=0.75\linewidth]{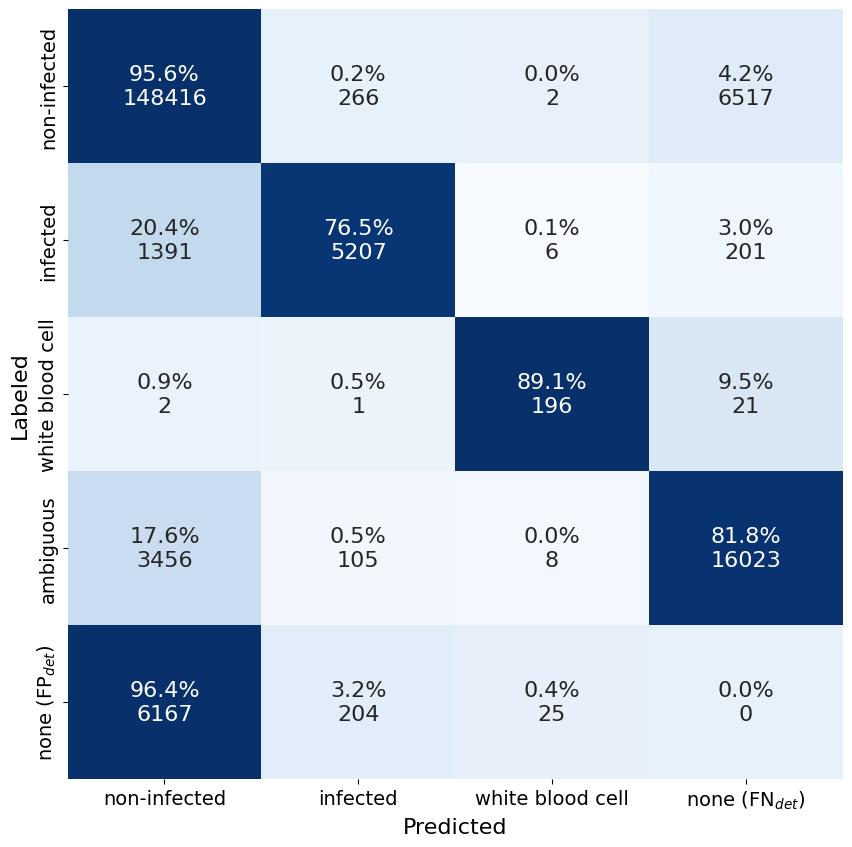}
\end{subfigure}
\begin{subfigure}[b]{\linewidth}
\centering
\caption{\label{fig:cm_poly} train: \mira, test: \polys}
\includegraphics[width=0.75\linewidth]{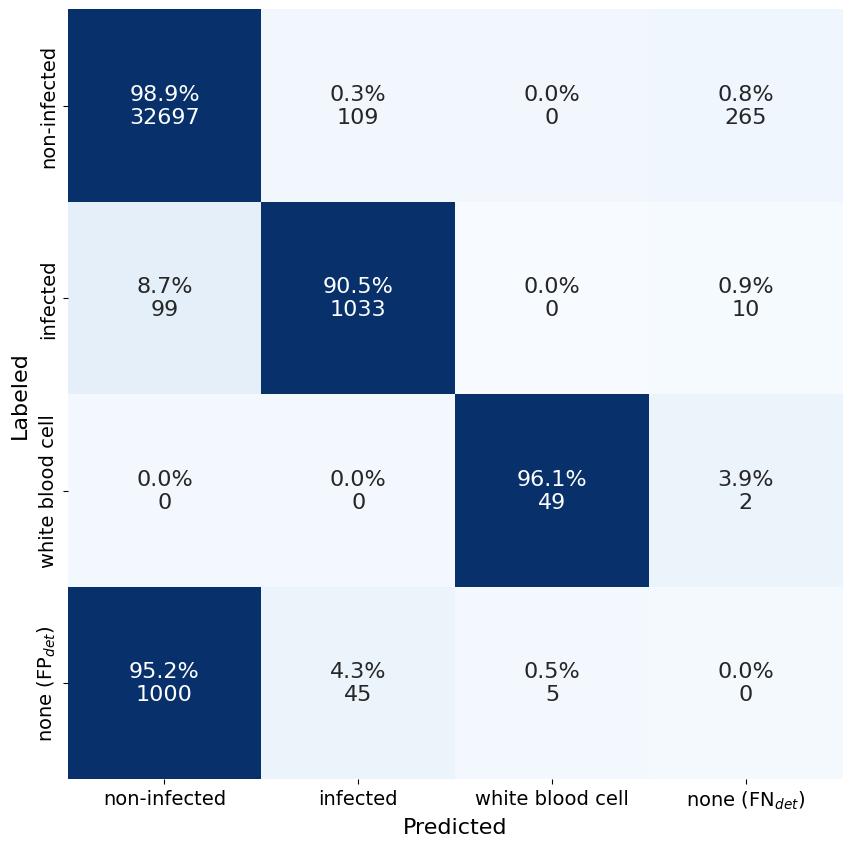}
\end{subfigure}%
\caption{\label{fig:cms} Confusion matrices for Faster R-CNN predictions on the NIH subsets. Each matrix shows row-normalized percentages along with absolute cell counts. The last row indicates \acfp{fp}, \ie, cell instances detected by the model but not annotated in the dataset. The last column indicates \acfp{fn}, \ie, annotated cell instances that were not detected by the model.}
\end{figure}

\Cref{fig:cm_points} presents the confusion matrix of the Faster R-CNN model trained on the \polys subset and evaluated on the \mira subset, and vice versa. Results are displayed as row-normalized percentages along with absolute cell counts. 

\begin{figure*}[t]
\centering
\begin{subfigure}[b]{0.49\linewidth}
\centering
\caption{\label{fig:non-poly} Sample from \polys}
\includegraphics[width=\linewidth]{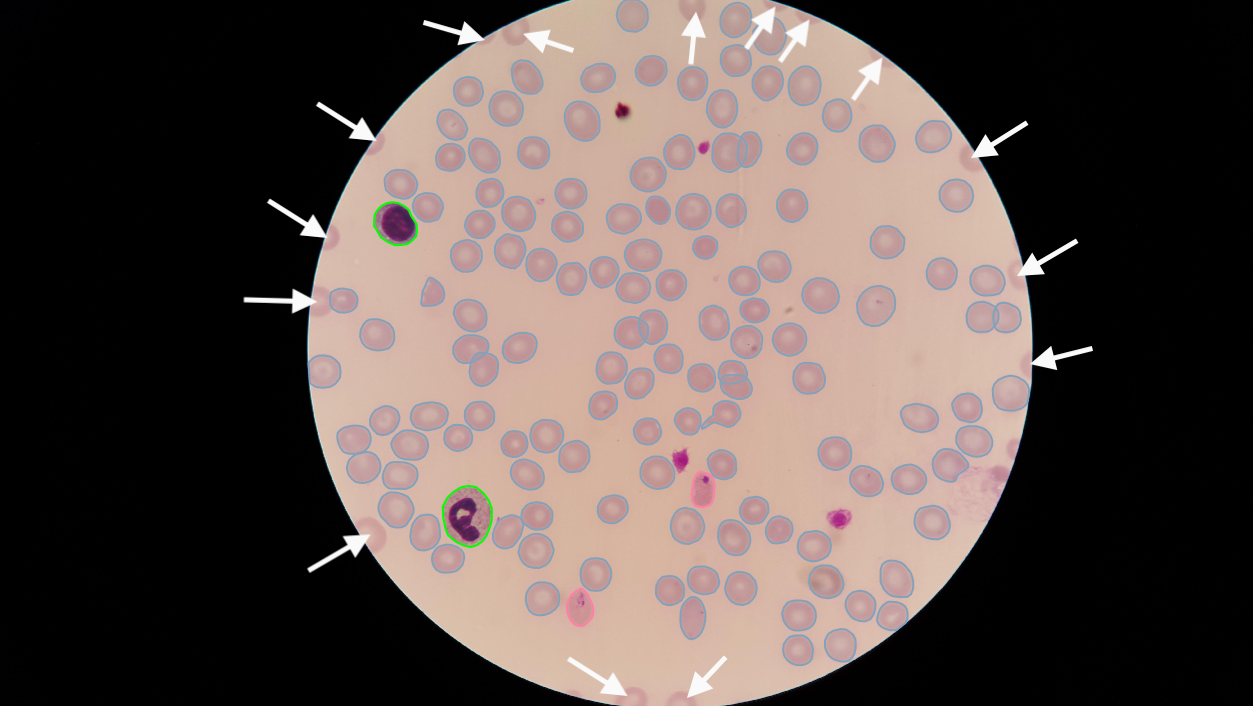}
\end{subfigure}\hfill
\begin{subfigure}[b]{0.49\linewidth}
\centering
\caption{\label{fig:non-points} Sample from \points}
\includegraphics[width=\linewidth]{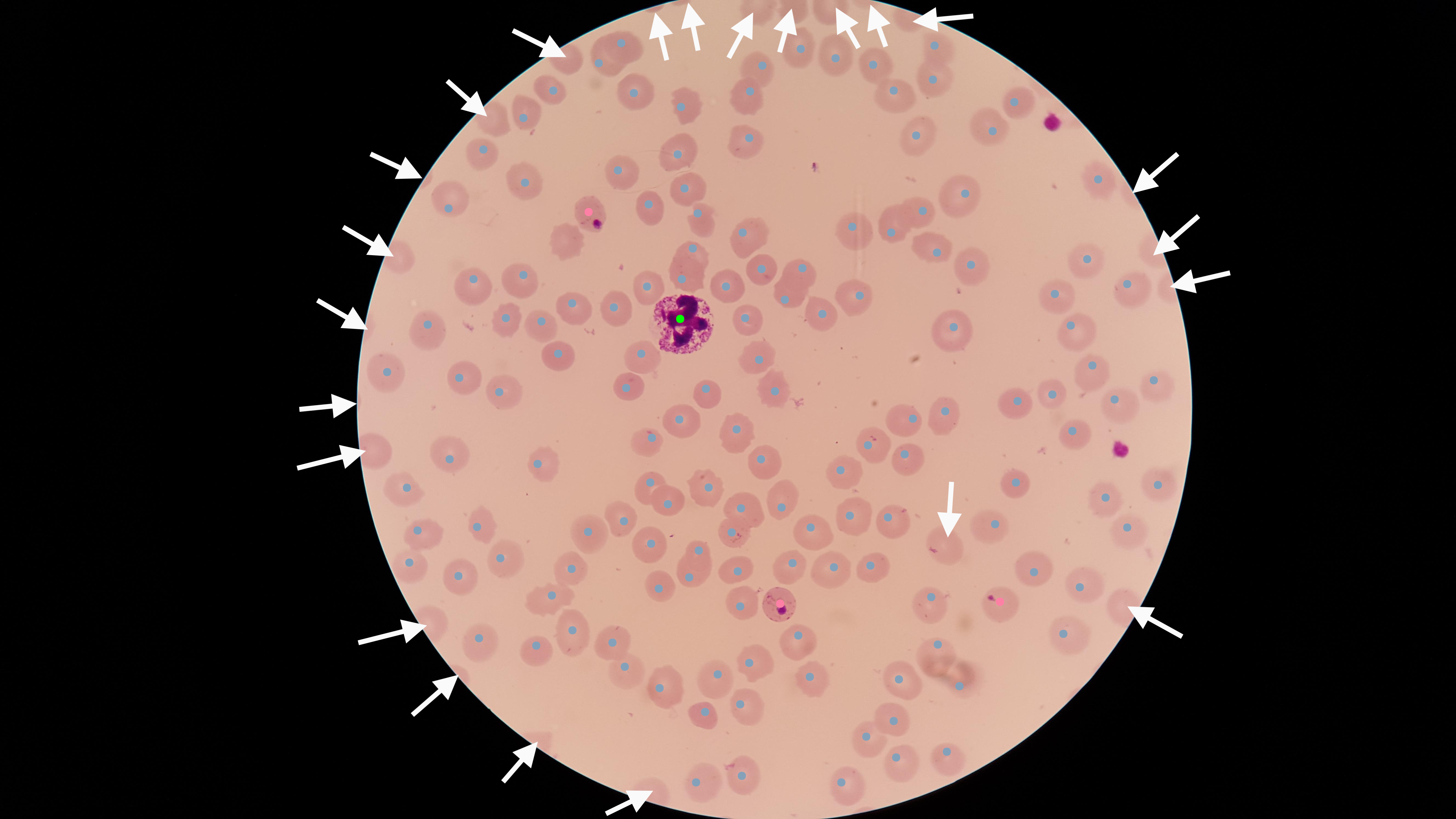}
\end{subfigure}%
\caption{\label{fig:non-anno} Representative samples from NIH subsets with white arrows indicating non-annotated cells at the border of field of view: (a) sample from the polygon subset with detailed contour annotations, (b) sample from the point subset with spot annotations in the cell center.}
\end{figure*}

\noindent Overall, the model performs better when trained on the \mira and evaluated on the \polys subset than the other way round, reflected by a lower proportion of off-diagonal entries in the confusion matrix. When training on \polys and testing on \mira, a comparably high ratio ($>20\,\%$) of infected cells was misclassified as non-infected, indicating reduced recall for malaria detection. Furthermore, \SI{81.8}{\percent} of the cells annotated as \emph{ambiguous} were not detected by the model. Closer inspection of these cells revealed that ambiguous cells were often located near the field-of-view borders, where annotations were inconsistently applied. Specifically, these border cells were frequently unannotated in both, the \polys and \points dataset. This suggests a possible labeling bias, which is further illustrated in \cref{fig:non-anno}, where white arrows indicate unlabeled yet clearly visible cells.

\Cref{tab:results} summarizes the detection performance, reported as precision, recall, and F1\,scores computed from the confusion matrices according to \cref{eq:f1,eq:prec,eq:rec}. For each dataset, training was repeated with three different random seeds, and we report the average performance as mean $\pm$ standard deviation ($\mu \pm \sigma$). 

\begin{table}[ht]
\centering
\caption{\label{tab:results} Class-wise F1\,score ($\mu \pm \sigma$) of detection model trained on \polys subset and evaluated on the \mira subset and vice versa.}
\resizebox{\linewidth}{!}{
\begin{tabular}{lS[table-format=1.2(2)]S[table-format=1.2(2)]}
&  \text{\polys $\rightarrow$ \mira} & \text{\mira $\rightarrow$ \polys}\\
\hline
\textbf{Precision} \\
\hspace{.25cm} non-infected cells & 0.96 \pm 0.01 &  0.97 \pm 0.00 \\
\hspace{.25cm} infected cells & 0.91 \pm 0.01 &  0.86 \pm 0.01 \\
\hspace{.25cm} white blood cells & 0.90 \pm 0.04 & 0.88 \pm 0.03 \\
\textbf{Recall} \\
\hspace{.25cm} non-infected cells & 0.97 \pm 0.01 &  0.99 \pm 0.00 \\
\hspace{.25cm} infected cells & 0.77 \pm 0.01 & 0.91 \pm 0.01 \\
\hspace{.25cm} white blood cells & 0.92 \pm 0.03 & 0.96 \pm 0.00 \\
\textbf{F1\,score} \\
\hspace{.25cm} non-infected cells & 0.96 \pm 0.01 & 0.98 \pm 0.00 \\
\hspace{.25cm} infected cells & 0.84 \pm 0.01 & 0.88 \pm 0.00 \\
\hspace{.25cm} white blood cells & 0.91 \pm 0.04 & 0.92 \pm 0.02 \\
\end{tabular}}
\end{table}

The results demonstrate high performance for the detection of non-infected red blood cells and white blood cells, with F1 scores above \SI{90}{\percent}.  Infected cells were detected with an average F1 score of \num{0.84}, when training on \polys and \num{0.88} when training on \mira, indicating good but comparatively lower performance. The repeated training runs demonstrate low variability, indicated by a low standard deviation of performance results.  

The performance metrics again demonstrate a superior performance of the model trained on \mira. This especially holds for the recall of infected cells, with average values of \num{0.77} for training on \polys and  \num{0.91} for training on \mira. This observation could be attributed to discrepancies in labeling consistency, but also to the higher volume of annotated instances (\num{6810} vs. \num{1142}), which provides more diverse training examples to the model.

\section{Discussion and Summary}
This study presents a revised version of the \acs{nih} malaria dataset with instance-level annotations in COCO format, facilitating the development of deep learning-based object detection models for the automatic detection of infected cells. We validated these annotations by training a \rcnn to detect infected and non-infected red blood cells, as well as white blood cells, achieving an F1\,score of up to \num{0.88} for the detection of infected cells. For trained microscopists, the \ac{who} guidelines recommend a minimum recall of infected malaria samples of \num{0.90}~\citep{who}, which our system achieves on a cellular level. Therefore, the system meets the minimum competency level required in a diagnostic setting. Nevertheless, our analysis of ambiguous cells revealed inconsistencies in the original annotations, where especially at the image borders cells were not labeled by the pathologists. However, it is difficult to tell whether these cells were simply overlooked or not labeled on purpose as a reliable malaria diagnosis might not be possible on partially visible cells. This raises broader concerns about ground truth quality in biomedical datasets likely caused by a trade-off of labeling precision and time investment. To the best of our knowledge, the original dataset was annotated by a single expert, which can introduce a considerable labeling bias. Future work could address this with additional manual annotation rounds with consensus labeling by multiple experts or introducing a separate class for partially visible cells. For evaluating the performance of machine learning models, we recommend excluding these cells from evaluation. 

Despite the challenges associated with partially labeled data, our results demonstrate that annotation conversion via existing tools such as Cellpose, followed by targeted manual curation, can yield training data of sufficient quality to support robust model performance. This finding is particularly relevant for resource-constrained settings where detailed annotations are expensive or infeasible.

In addition to annotation consistency, we also observed differences in model performance between the two subsets of the \ac{nih} dataset, likely driven by the varying number of annotated instances available for training. This highlights the importance of dataset size and diversity for learning subtle morphological features, such as the presence of ring-stage parasites. Furthermore, our initial dataset assessment demonstrated a high class imbalance of healthy and infected cells. We compensated for this to some extent by employing a customized patch sampling strategy, but in future work dedicated augmentation strategies or class-balanced loss functions could be integrated.   

Overall, our work contributes an enhanced dataset and a robust pipeline for parasite detection in microscopy, supporting further research into automated malaria diagnosis.

\coi{The authors do not have any conflicts of interest to declare.}

\acks{The authors acknowledge the U.S. National Library of Medicine for making the thin blood smear dataset used in this work publicly available. The dataset is provided under a license that permits redistribution and modification, with appropriate attribution. The authors gratefully acknowledge the scientific support and HPC resources provided by the Erlangen National High Performance Computing Center (NHR@FAU) of the Friedrich-Alexander-Universität Erlangen-Nürnberg (FAU). The hardware is funded by the German Research Foundation (DFG).}

%
\ethics{The work follows appropriate ethical standards in conducting research and writing the manuscript, in accordance with all applicable laws and regulations regarding the treatment of human subjects. The dataset used in this study was publicly released by \citet{kassim2020clustering}, who obtained the necessary ethical approvals as documented in the original publication. No additional ethical approval was required.}

\data{The dataset~\citep{kassim2020clustering} used in this study was developed and funded by the U.S. \acf{nlm}, part of the \acf{nih}, and is publicly available for commercial and non-commercial use. Use of this dataset is governed by a license that requires proper attribution. We acknowledge the source of the data as follows: ``Courtesy of the U.S. National Library of Medicine.'' The dataset and associated information are available at: \url{https://lhncbc.nlm.nih.gov/publication/pub9932}. Please cite the dataset as described by \citet{kassim2020clustering}. The updated annotation set with the modifications described in \cref{sec:annos} is publicly available via Zenodo: \url{https://doi.org/10.5281/zenodo.17514694}.}

\bibliography{literature}

\begin{acronym}
\acro{nih}[NIH]{National Institutes of Health}
\acro{nlm}[NML]{National Library of Medicine}
\acro{map}[mAP]{mean average precision}
\acro{nms}[NMS]{non-maximum suppression}
\acro{fn}[FN]{false negative}
\acro{fp}[FP]{false positive}
\acro{fn-det}[$FN_{det}$]{false negatives due to detection failure}
\acro{fp-det}[$FP_{det}$]{false positives due to detection failure}
\acro{who}[WHO]{World Health Organization }
\end{acronym}

\end{document}